\documentclass[runningheads,a4paper]{llncs}
\usepackage{amsfonts} % assumes amsmath package installed
\usepackage{amsmath} 
\usepackage{amssymb}
\usepackage{array}
\usepackage{bm}
\usepackage{graphics} % for pdf, bitmapped graphics files
\usepackage{graphicx} % more modern
\usepackage{hyperref}
\usepackage{vector}

\renewcommand{\bvec}[1]{\ensuremath{\bm{#1}}}
\usepackage{url}

\begin{document}

\mainmatter  % start of an individual contribution

% first the title is needed
\title{Efficient Baseline-free Sampling in Parameter Exploring Policy Gradients:\\Super Symmetric PGPE}

% a short form should be given in case it is too long for the running head
\titlerunning{Super Symmetric PGPE}

\author{Frank Sehnke}
\authorrunning{Frank Sehnke}
\institute{Zentrum f{\"u}r Sonnenenergie- und Wasserstoff-Forschung,\\
Industriestr. 6, Stuttgart, BW 70565 Germany}

\maketitle

\begin{abstract}
Policy Gradient methods that explore directly in parameter space are among the most effective and robust direct policy
search methods and have drawn a lot of attention lately. 
The basic method from this field, Policy Gradients with Parameter-based Exploration, uses two samples that are 
symmetric around the current hypothesis to circumvent misleading reward in \emph{asymmetrical} reward distributed 
problems gathered with the usual baseline approach.
The exploration parameters are still updated by a baseline approach - leaving the exploration prone to asymmetric 
reward distributions.
In this paper we will show how the exploration parameters can be sampled quasi symmetric despite having limited instead
of free parameters for exploration.
We give a transformation approximation to get quasi symmetric samples with respect to the exploration without changing
the overall sampling distribution.
Finally we will demonstrate that sampling symmetrically also for the exploration parameters is 
superior in needs of samples and robustness than the original sampling approach.
\end{abstract}
%
%=====================================================================================================================%
%                                        ============== Section =============                                         %
%                                        =========== Introduction ===========                                         %
%                                        ========== sec:introduction ========                                         %
%=====================================================================================================================%
\section{Introduction}\label{sec:indtroduction}
Policy Gradient (PG) methods that explore directly in parameter space have some major advantages over standard PG 
methods, like described in \cite{sehnke2010parameter,ruckstiess2010exploring,Miyamae2010,Zhao2011a,zhao2013efficient,stulp2012path}
and \cite{wierstra2008natural} and have therefore drawn a lot of attention in the last years. 
The basic method from the field of Parameter Exploring Policy Gradients (PEPG) \cite{sehnke2012parameter}, Policy 
Gradients with Parameter-based Exploration (PGPE) \cite{sehnke2010parameter}, uses two samples that are symmetric 
around the current hypothesis to circumvent misleading reward in \emph{asymmetrical} reward distributed problems, 
gathered with the usual baseline approach. 
\cite{Zhao2011a} shows that Symmetric Sampling (SyS) is superior even to the optimal baseline.
The exploration parameters, however, are still updated by a baseline approach - leaving the exploration prone to 
asymmetric reward distributions. 
While the optimal baseline improved this issue substantially, like shown again by \cite{Zhao2011a}, it is likely 
that removing the baseline altogether by a SyS wrt. the exploration parameters will be again superior. 
Because the exploration parameters are standard deviations that are bounded between zero and infinity, there exist no 
correct symmetric samples wrt. the exploration parameters.
We will, however, show how the exploration parameters can be sampled quasi symmetric.
We give therefore a transformation approximation to get quasi symmetric samples without changing the 
overall sampling distribution significantly, so that the PGPE assumptions based on normal distributed samples still 
hold.
Finally we will demonstrate via experiments that sampling symmetrically also for the exploration parameters is superior
in needs of samples and robustness compared to the original sampling approach, if confronted with search spaces with 
significant amounts of local optima.
%=====================================================================================================================%
%                                        ============== Section =============                                         %
%                                        =========== Introduction ===========                                         %
%                                        ============ sec:method ============                                         %
%=====================================================================================================================%
\section{Method}\label{sec:method}
In this section we derive the super-symmetric sampling (SupSyS) method. We show how the method relates to SyS and
sampling with a baseline, thereby summarizing the derivation from \cite{sehnke2010parameter} for SyS and baseline sampling PGPE.
%=====================================================================================================================%
%===                           Parameter-Based Exploration === \label{subsec:paraBasedExplo}                       ===%
%=====================================================================================================================%
\subsection{Parameter-Based Exploration}\label{subsec:paraBasedExplo}
To stay conform with the nomenclature of \cite{sehnke2010parameter} and \cite{Zhao2011a}, we assume a Markovian environment that 
produces a cumulative reward $r$ for a fixed length \emph{episode}, \emph{history}, \emph{trajectory} or \emph{roll-out}.  
In this setting, the goal of reinforcement learning is to find the optimal policy parameters $\bvec{\theta}$ that 
maximize the agent's expected reward
\begin{equation}
\label{eqn:defJ}
J(\bvec{\theta}) = \int_H p(h|\bvec{\theta}) r(h) dh.
\end{equation}
An obvious way to maximize $J(\bvec{\theta})$ is to estimate $\nabla_{\bvec{\theta}} J$ and use it to carry out gradient ascent 
optimization. 
The probabilistic policy used in standard PG is replaced with a probability distribution over the parameters $\bvec{\theta}$ 
for PGPE.
The advantage of this approach is that the actions are deterministic, and an entire history can therefore be generated 
from a single parameter sample. 
This reduction in samples-per-history is what reduces the variance in the gradient estimate (see~\cite{sehnke2010parameter} for 
details). 

We name the distribution over parameters in accordance with \cite{sehnke2010parameter} $\bvec{\rho}$.
The expected reward with a given $\bvec{\rho}$ is
\begin{equation}
\label{eqn:coolJ}
J(\bvec{\rho}) = \int_{\bvec{\Theta}} \int_H p(h, \bvec{\theta}|\bvec{\rho}) r(h) dh d\bvec{\theta}.
\end{equation}
Differentiating this form of the expected return with respect to $\bvec{\rho}$ and applying sampling methods (first choosing $\bvec{\theta}$ from $p(\bvec{\theta}|\bvec{\rho})$, then running the agent to generate $h$ from $p(h|\bvec{\theta})$) yields the following gradient estimator:
\begin{equation}
\label{eqn:approx_pgpe}
\nabla_{\bvec{\rho}} J(\bvec{\rho}) \approx \frac{1}{N}\sum_{n=1}^{N}{\nabla_{\bvec{\rho}} \log p(\bvec{\theta}|\bvec{\rho})r(h^n)}.
\end{equation}
Assuming that $\bvec{\rho}$ consists of a set of means $\{\mu_i\}$ and standard deviations $\{\sigma_i\}$ that determine an independent normal distribution for each parameter $\theta_i$ in $\bvec{\theta}$ 
gives the following forms for the derivative of the characteristic eligibility $\log p(\bvec{\theta}|\bvec{\rho})$ with respect to $\mu_i$ and $\sigma_i$
\begin{equation}
  \label{eqn:updates}
  \nabla_{\mu_i} \log p(\bvec{\theta} | \bvec{\rho}) = \frac{(\theta_i -\mu_i)}{\sigma_i^2}, \,\,\,\,\,\,\,\,\,\,
  \nabla_{\sigma_i} \log p(\bvec{\theta}| \bvec{\rho}) = \frac{(\theta_i - \mu_i)^{2} - \sigma_i^2}{\sigma_i^{3}},
\end{equation}
which can be substituted into Eq.~\eqref{eqn:approx_pgpe} to approximate the $\bvec{\mu}$ and $\bvec{\sigma}$ gradients.
%=====================================================================================================================%
%===                                Sampling with a Baseline === \label{subsec:baseline}                           ===%
%=====================================================================================================================%
\subsection{Sampling with a Baseline}\label{subsec:baseline}
Given enough samples, Eq.~\eqref{eqn:approx_pgpe} will determine the reward gradient to arbitrary accuracy. 
However each sample requires rolling out an entire state-action history, which is expensive. 
Following \cite{Williams:92}, we obtain a cheaper gradient estimate by drawing a single sample $\bvec{\theta}$ and comparing its reward $r$ to a baseline reward $b$ given e.g. by a moving average over previous samples. 
Intuitively, if $r > b$ we adjust $\bvec{\rho}$ so as to increase the probability of $\bvec{\theta}$, and $r < b$ we do the opposite. 
If, as in \cite{Williams:92}, we use a step size $\alpha_i = \alpha \sigma_i^2$ in the direction of positive gradient (where $\alpha$ is a constant) we get the following parameter update equations:
\begin{equation}
  \label{eqn:frank1}
  \Delta \mu_i = \alpha (r-b)(\theta_i - \mu_i), \,\,\,\,\,\,\,\,\,\,
  \Delta \sigma_i = \alpha (r-b) \frac{(\theta_i - \mu_i)^{2} - \sigma_i^{2}}{\sigma_i}.
\end{equation}
Usually the baseline is realized as decaying or moving average baseline of the form:
\begin{equation}
  \label{eqn:avBase}
  b(n) = \gamma r(h^{n-1}) + (1-\gamma) b(n-1) \,\,\,\,\,\,\,\,\,\, \text{or} \,\,\,\,\,\,\,\,\,\,
  b(n) = \sum_{n=N-m}^{N}{r(h^n)}/m
\end{equation}
\cite{Zhao2011a} showed recently that an optimal baseline can be achieved for PGPE and the algorithm converges significantly faster with an optimal baseline of the form:
\begin{equation}
  \label{eqn:opBase}
  b^* = \frac{\mathbb{E}[r(h) || \nabla_{\bvec{\rho}} \log p(\bvec{\theta} | \bvec{\rho}) ||^2]}{\mathbb{E}[|| \nabla_{\bvec{\rho}} \log p(\bvec{\theta} | \bvec{\rho}) ||^2]}.
\end{equation}

%=====================================================================================================================%
%===                                   Symmetric Sampling === \label{subsec:sys}                                   ===%
%=====================================================================================================================%
\subsection{Symmetric Sampling}\label{subsec:sys}
While sampling with a baseline is efficient and reasonably accurate for most scenarios, it has several drawbacks. 
In particular, if the reward distribution is strongly skewed then the comparison between the sample reward and the baseline reward is misleading. 
A more robust gradient approximation can be found by measuring the difference in reward between two symmetric samples on either side of the current mean. 
That is, we pick a perturbation $\bvec{\epsilon}$ from the distribution $\mathcal{N}(0,\bvec{\sigma})$, then create symmetric parameter samples $\bvec{\theta}^+ = \bvec{\mu} + \bvec{\epsilon}$ and $\bvec{\theta}^- = \bvec{\mu} - \bvec{\epsilon}$. 
Defining $r^+$ as the reward given by $\bvec{\theta}^+$ and $r^-$ as the reward given by $\bvec{\theta}^-$. 
We can insert the two samples into Eq.~\eqref{eqn:approx_pgpe} and make use of Eq.~\eqref{eqn:updates} to obtain 
\begin{equation}
\nabla_{\mu_i} J(\bvec{\rho}) \approx \frac{\epsilon_i(r^+ - r^-)}{2 \sigma_i^2},
\end{equation}
which resembles the \emph{central difference} approximation used in finite difference methods. 
Using the same step sizes as before gives the following update equation for the $\bvec{\mu}$ terms
\begin{equation}
  \label{eqn:sysMuUpdate}
\Delta \mu_i = \frac{\alpha\epsilon_i(r^+ - r^-)}{2}.
\end{equation}
The updates for the standard deviations are more involved. 
As $\bvec{\theta}^+$ and $\bvec{\theta}^-$ are by construction equally probable under a given $\bvec{\sigma}$, the difference between them cannot be used to estimate the $\bvec{\sigma}$ gradient. 
Instead we take the mean $\frac{r^+ +\, r^-}{2}$ of the two rewards and compare it to the baseline reward $b$. This approach yields
\begin{equation}
 \label{eqn:sysSigUpdate}
\Delta \sigma_i = \alpha\left(\frac{r^+ + r^-}{2} - b\right)\left(\frac{\epsilon_i^2 - \sigma_i^2}{\sigma_i}\right)
\end{equation}
SyS removes the problem of misleading baselines, and therefore improves the $\bvec{\mu}$ gradient estimates. 
It also improves the $\bvec{\sigma}$ gradient estimates, since both samples are equally probable under the current distribution, and therefore reinforce each other as predictors of the benefits of altering $\bvec{\sigma}$. 
Even though symmetric sampling requires twice as many histories per update, \cite{sehnke2010parameter} and \cite{Zhao2011a} have shown that it gives a considerable improvement in convergence quality and time.
%=====================================================================================================================%
%===                             Super-Symmetric Sampling === \label{subsec:SupSyS}                              ===%
%=====================================================================================================================%
\subsection{Super-Symmetric Sampling}\label{subsec:SupSyS}
\begin{figure}[t]
\centering
\begin{minipage}[t]{0.485\textwidth}
  \centering
  \includegraphics[scale=0.3333]{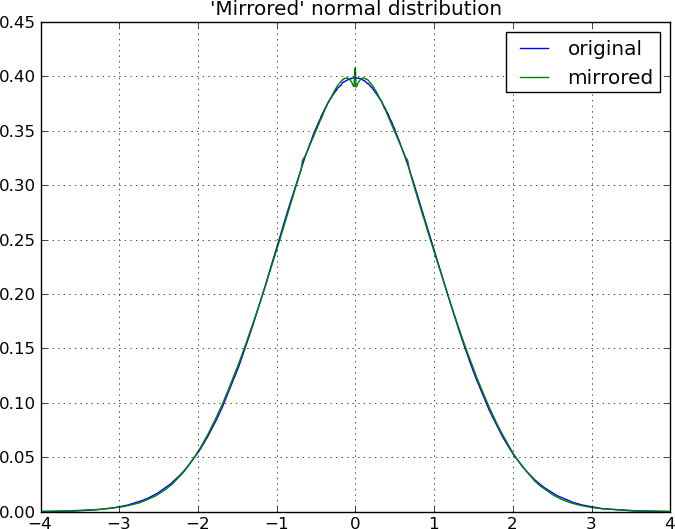}
  \caption{\label{fig:mirrorDistriSecond} Normal distribution and the final approximation of the 'mirrored' distribution.}
\end{minipage}
\hspace{0.01\textwidth}
\begin{minipage}[t]{0.485\textwidth}
  \centering
  \includegraphics[scale=0.3333]{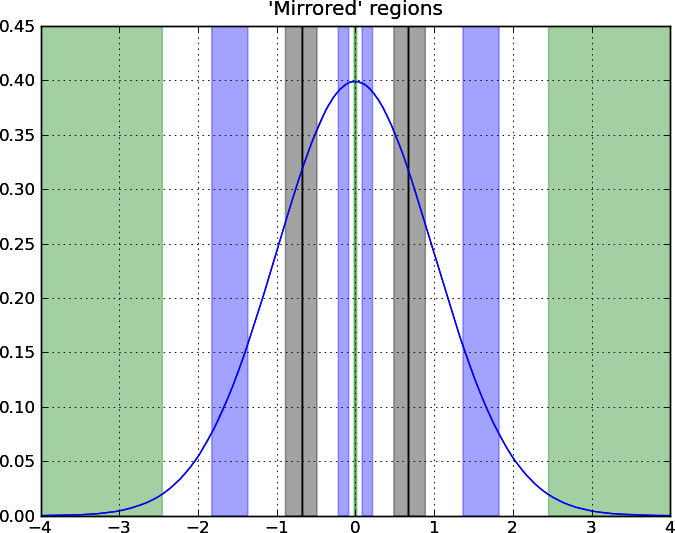}
  \caption{\label{fig:mirrorDistriRegions} Normal distribution and the regions that are transfered into each other by 'reflecting' the samples on the other side of the median deviation.}
\end{minipage}
\end{figure}
While SyS removes the misleading baseline problem for the $\bvec{\mu}$ gradient estimate, the $\bvec{\sigma}$ gradient still uses a baseline and is prone to this problem.
On the other hand there is no correct \emph{symmetric} sample with respect to the standard deviation, because the standard deviation is bounded on the one \emph{side} to $0$ and is unbounded on the positive \emph{side}.
Another problem is that $\frac{2}{3}$ of the samples are on one \emph{side} of the standard deviation and only $\frac{1}{3}$ on the other - \emph{mirroring} the samples to the opposite side of the standard deviation in some way, would therefore deform the normal distribution so much, that it would no longer be a close enough approximation to fulfill the assumptions that lead to the PGPE update rules.

We therefore chose to define the normal distribution via the mean and the median deviation $\bvec{\phi}$.
The median deviation is due to the nice properties of the normal distribution simply defined by:  $\bvec{\phi} = 0.67449 \cdot \bvec{\sigma}$.
We can therefore draw samples from the new defined normal distribution: $\bvec{\epsilon} \sim \mathcal{N}_m(0,\bvec{\phi})$.

The median deviation has by construction an equal amount of samples on either side and solves therefore the symmetry problem of \emph{mirroring} samples.
The update rule Eq.~\eqref{eqn:sysMuUpdate} stays unchanged while Eq.~\eqref{eqn:sysSigUpdate} is only scaled by $\frac{1}{0.67449}$ (the factor that transforms $\bvec{\phi}$ in $\bvec{\sigma}$) that can be substituted in $\alpha_{\bvec{\sigma}}$.

While the update rules stay the same for normal distributed sampling using the median deviation (despite a larger $\alpha_{\bvec{\sigma}}$), the median deviation is still also bounded on one side. 
Because the \emph{mirroring} cannot be solved in closed form we resort to approximation via a polynomial that can be transfered to an infinite series. 
We found a good approximation for \emph{mirroring} samples by:
\begin{equation}
  a_i = \frac{\phi_i - \mid \epsilon_i \mid}{\phi_i},\,\,\,\,\,\,\,\,\,\, 
  \epsilon_i^* = sign(\epsilon_i) \cdot \phi_i \cdot \begin{cases} 
    e^{c_1 \frac{{\mid a_i \mid}^3-\mid a_i \mid}{\log(\mid a_i \mid)} + c_2 \mid a_i \mid}
    &\mbox{if } a_i \leq 0 \\

    e^{a_i}/(1.-a_i^3)^{c_3 a_i}
    &\mbox{if } a_i > 0, 
  \end{cases} 
\end{equation}
with the following constants:
$c_1 = -0.06655, c_2 = -0.9706, c_3 = 0.124.$
This \emph{mirrored} distribution has a standard deviation of 1.002 times the original standard deviation and looks like depicted in Fig.~\ref{fig:mirrorDistriSecond}.
Fig.~\ref{fig:mirrorDistriRegions} shows the regions of samples that are transfered into each other while generating the quasi symmetric samples.

Additional to the symmetric sample with respect to the mean hypothesis, now we also can generate two quasi symmetric samples with respect to the median deviation.
We named this set of four samples super symmetric samples (SupSyS-samples). They allow for completely baseline free update rules, not only for the $\bvec{\mu}$ update but also for the $\bvec{\sigma}$ updates.

Therefore the two symmetric sample pairs are used to update $\bvec{\mu}$ according to Eq.~\eqref{eqn:sysMuUpdate}.
$\bvec{\sigma}$ is updated in a similar way by using the mean reward of each symmetric sample pair, there $r^{++}$ is the mean reward of the original symmetric sample pair and $r^{--}$ is the mean reward of the \emph{mirrored} sample pair.
The SupSyS update rule for the $\bvec{\sigma}$ update is given by:
\begin{equation}
  \label{eqn:supSysSigUpdate}
\Delta \sigma_i = \frac{\alpha\frac{\epsilon_i^2 - \sigma_i^2}{\sigma_i}(r^{++} - r^{--})}{2}.
\end{equation}
%
%=====================================================================================================================%
%                                        ============== Section =============                                         %
%                                        ============ Experiments ===========                                         %
%                                        ============ section:exp ===========                                         %
%=====================================================================================================================%
\section{Experiments and Results}\label{sec:exp}
\begin{figure}[t]
\centering
\begin{minipage}[t]{0.485\textwidth}
  \centering
  \includegraphics[scale=0.3]{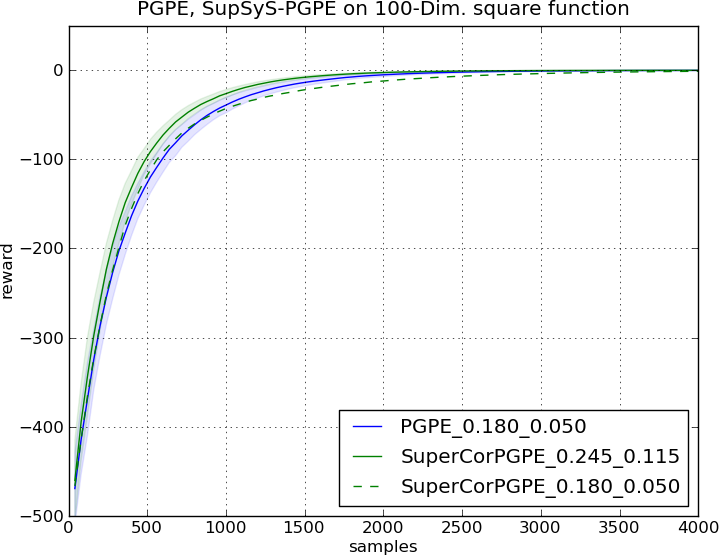}
  \caption{\label{fig:sqare100} Convergence plots of PGPE and SupSyS-PGPE on the 100 dimensional square function. The mean and standard deviation of 200 independent runs are shown.}
\end{minipage}
\hspace{0.01\textwidth}
\begin{minipage}[t]{0.485\textwidth}
  \centering
  \includegraphics[scale=0.3]{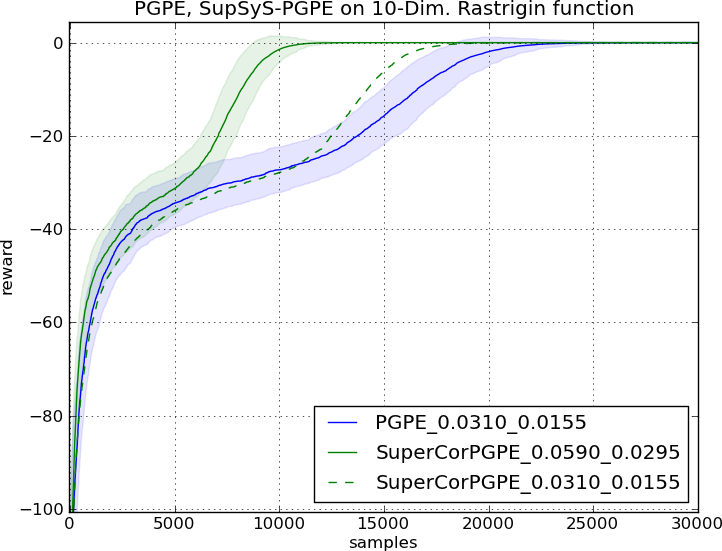}
  \caption{\label{fig:rast10} Convergence plots of PGPE and SupSyS-PGPE on the 10 dimensional Rastrigin function. The mean and standard deviation of 200 independent runs are shown.}
\end{minipage}
\end{figure}
\begin{figure}[b!]
\centering
\begin{minipage}[t]{0.485\textwidth}
  \centering
  \includegraphics[scale=0.3]{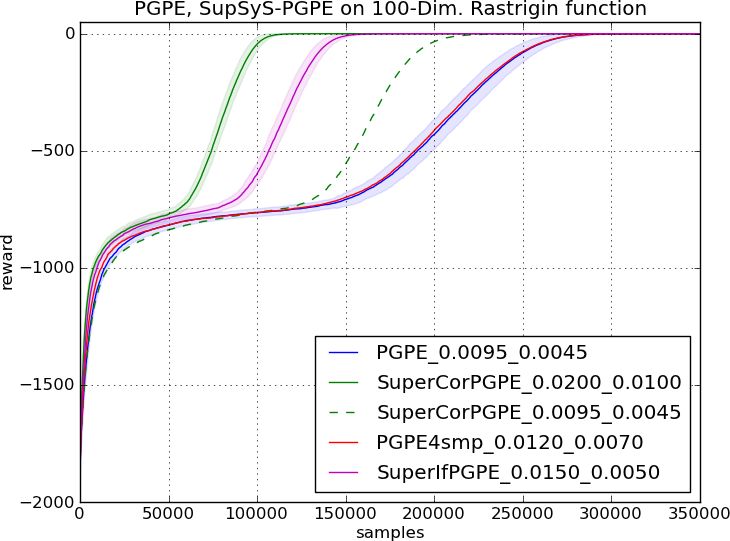}
  \caption{\label{fig:rast100} Convergence plots of PGPE, PGPE with 4 samples (PGPE4smp), conditional SupSyS-PGPE (SupIf-PGPE) and SupSyS-PGPE on the 100 dimensional Rastrigin function. The mean and standard deviation of 200 independent runs are shown.}
\end{minipage}
\hspace{0.01\textwidth}
\begin{minipage}[t]{0.485\textwidth}
  \centering
  \includegraphics[scale=0.3]{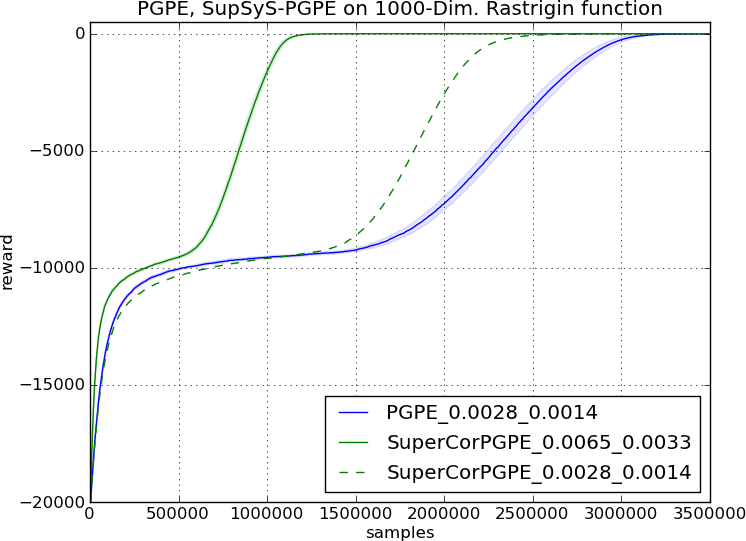}
  \caption{\label{fig:rast1000} Convergence plots of PGPE and SupSyS-PGPE on the 1000 dimensional Rastrigin function. The mean and standard deviation of 200 independent runs are shown.}
\end{minipage}
\end{figure}
\begin{figure}[t]
\centering
\begin{minipage}[t]{0.485\textwidth}
  \centering
  \includegraphics[scale=0.3]{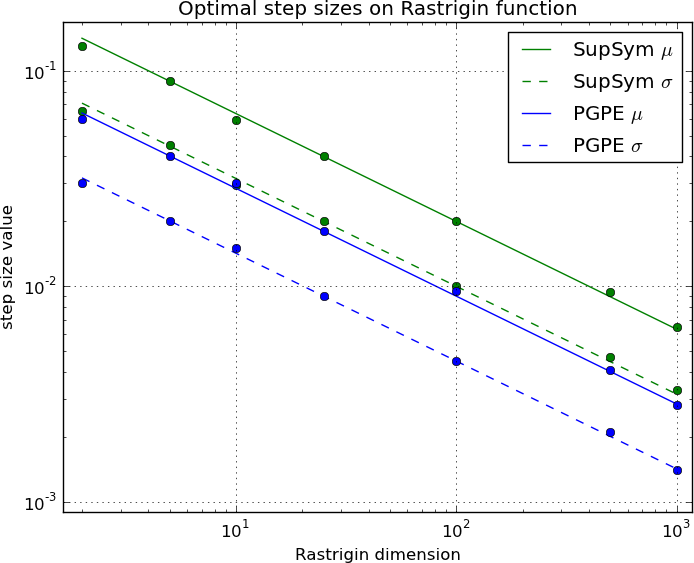}
  \caption{\label{fig:rastOver} Optimal meta-parameters for the multi-dimensional Rastrigin function for PGPE and SupSyS-PGPE.}
\end{minipage}
\hspace{0.01\textwidth}
\begin{minipage}[t]{0.485\textwidth}
  \centering
  \includegraphics[scale=0.3333]{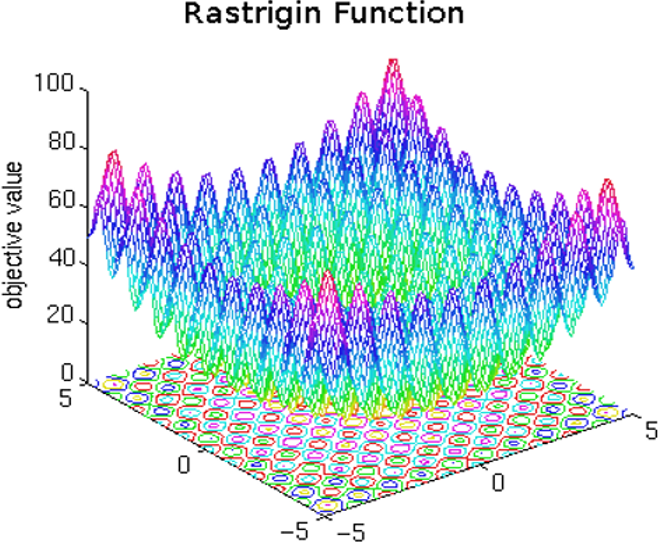}
  \caption{\label{fig:rast2D} Visualization of the 2D Rastrigin function.}
\end{minipage}
\end{figure}
We use the square function as search space instance with no local optima and the Rastrigin function (see Fig.~\ref{fig:rast2D}) as search space with exponentially many local optima, to test the different behavior of SupSyS- and SyS-PGPE.
The two meta-parameters connected with SyS-PGPE as well as with SupSyS-PGPE, namely the step sizes for the $\bvec{\mu}$ and $\bvec{\sigma}$ updates, were optimized for every experiment via a grid search. The Figures~\ref{fig:sqare100} to \ref{fig:rast1000} show the means and standard deviations of 200 independent runs each.
It can be seen in Fig.~\ref{fig:sqare100} that for a search space with no local optima SupSyS-PGPE shows no advantage over standard SyS-PGPE.
However, despite using 4 samples per update the performance is also not reduced by using SupSyS-PGPE --- the two methods become merely equivalent.
The situation changes drastically if the Rastrigin function is used as test function. Not only needs SupSyS-PGPE about half the samples compared to PGPE, the effect seems also to become stronger the higher dimensional the search space gets (see Fig.~\ref{fig:rast10} to Fig.~\ref{fig:rast1000}).
We also added SupSyS-PGPE plots with the for SyS-PGPE optimal (less greedy) meta parameters to show that the effect is not only due to the more aggressive meta parameters.
This runs were also more efficient than for PGPE, the effect was however not so distinct.

In Fig.~\ref{fig:rast100} we also show a standard PGPE experiment with 4 samples (2 SyS samples --- \emph{PGPE4smp}) instead of 2 to show that the improved performance is not due to the different samples per update. Fig.~\ref{fig:rast100} additionally shows an experiment (SupIf-PGPE) there symmetric samples are only drawn if the first sample(s) result in worse reward than a decaying average baseline. The intuitive idea behind symmetric samples was initially that changing the parameters \emph{away} from the current sample if the sample resulted in lower than average reward may move the mean hypothesis still in a worse region of the parameter space. Search spaces like the one given in the Rastrigin function can visualize this problem.
For SupIf-PGPE one Sample is drawn. If the reward is larger than the baseline then an update is done immediately. 
If not, a symmetric sample is drawn. Is the mean reward connected with both samples better than the baseline an SyS-PGPE update is done. 
If also this mean reward is worse than the baseline, a full SupSyS-PGPE update with 2 additional SyS samples is performed.
As can be seen in Fig.~\ref{fig:rast100} the performance is worse by some degree --- the difference is however small enough that maybe the optimal baseline approach would improve this method enough to be challenging to SupSyS-PGPE (see also Sec.~\ref{sec:future}).

The optimal meta-parameters are an exponential function of the search space dimension, like to expect, so that we observe a line in the \emph{loglog}-plot of Fig.~\ref{fig:rastOver}.
For SupSyS-PGPE the meta-parameters are about 2 times larger than for SyS-PGPE. This is partly because SupSyS-PGPE uses four samples per update instead of two. But the optimal meta-parameters are also larger than for the PGPE4smp experiment so that the symmetric nature of the four SupSyS samples obviously brings additional stability in the gradient estimate than a pure averaging over 4 samples would.
%=====================================================================================================================%
%                                        ============== Section =============                                         %
%                                        =========== Introduction ===========                                         %
%                                        =========== section:intro ==========                                         %
%=====================================================================================================================%
\section{Conclusions and Future Work}\label{sec:future}
We introduced SupSyS-PGPE, a completely baseline free PGPE that uses quasi-symmetric samples wrt. the exploration parameters.
We showed that on the Rastrigin function, as example for a test function with exponentially many local optima, this novel method is clearly superior to standard SyS-PGPE and that both methods become equivalent in performance if the search space lack \emph{distracting} local optima.

For future work we want to highlight that SupSyS-PGPE can be easily combined with other extensions of PGPE. Multi-modal PGPE \cite{sehnke2010multimodal} can be equipped straight forward with SupSyS sampling. Also the natural gradient used for PGPE in \cite{Miyamae2010} can be defined over the SupSyS gradient instead over the vanilla gradient. If the full 4 super symmetric sample set is only used if the first samples are worse than a baseline (like described as SupIf-PGPE in Sec.~\ref{sec:exp}) a combination with the optimal baseline (described for PGPE in \cite{Zhao2011a}) can yield a superior method to both SupSyS-PGPE and optimal baseline PGPE.
Also importance mixing introduced for PGPE by~\cite{zhao2013efficient} is applicable to SupSyS-PGPE.

Finally a big open point for future work is the validation of the mere theoretical findings on real world problems, e.g. robotic tasks, for SupSyS-PGPE and its combination with other PGPE extensions.
%=====================================================================================================================%
%                                        ============== Section =============                                         %
%                                        =========== Introduction ===========                                         %
%                                        =========== section:intro ==========                                         %
%=====================================================================================================================%
%\begin{thebibliography}
\bibliography{references}
\bibliographystyle{splncs}
%\end{thebibliography}
\end{document}